\documentclass[conference]{IEEEtran}
\IEEEoverridecommandlockouts

\usepackage{cite}
\usepackage{amsmath,amssymb,amsfonts}
\usepackage{algorithmic}
\usepackage{graphicx}
\usepackage{subcaption}
\usepackage{textcomp}
\usepackage{xcolor}
\usepackage{booktabs}
\usepackage{multirow}
\usepackage[printonlyused]{acronym}
\usepackage{url}
\usepackage{siunitx}
\DeclareSIUnit{\pp}{pp}
\newacro{CNN}[CNN]{Convolutional Neural Network}
\newacro{DRL}[DRL]{Deep Reinforcement Learning}
\newacro{FF}[FF]{Feedforward}
\newacro{GAE}[GAE]{Generalized Advantage Estimation}
\newacro{GRU}[GRU]{Gated Recurrent Unit}
\newacro{MLP}[MLP]{Multi-Layer Perceptron}
\newacro{PPO}[PPO]{Proximal Policy Optimization}
\newacro{SG}[SG]{Subgoal}
\newacro{SIMD}[SIMD]{Single Instruction, Multiple Data}
\newacro{WFC}[WFC]{Wave Function Collapse}

\begin{document}

\title{Beyond Specialization: Robust Reinforcement Learning Navigation via Procedural Map Generators}

\author{
Christian Jestel\textsuperscript{1,2}, Nicolas Bach\textsuperscript{1,2}, Marvin Wiedemann\textsuperscript{1,2}, Jan Finke\textsuperscript{1,2}, and Peter Detzner\textsuperscript{3}
\thanks{\textsuperscript{1}C.\ Jestel, N.\ Bach, M.\ Wiedemann, and J.\ Finke are with the Fraunhofer Institute for Material Flow and Logistics IML, Dortmund, Germany. Email: \{christian.jestel, nicolas.bach, marvin.wiedemann, jan.finke\}@iml.fraunhofer.de}
\thanks{\textsuperscript{2}The Lamarr Institute for Machine Learning and Artificial Intelligence, Germany}
\thanks{\textsuperscript{3}P.\ Detzner is with the Westphalian University of Applied Sciences, Gelsenkirchen, Germany. This work was done while at the Fraunhofer Institute for Material Flow and Logistics IML, Dortmund, Germany}
}

\maketitle

\begin{abstract}
\ac{DRL} navigation policies often overfit to the structure of their training environments, as environmental diversity is typically constrained by the manual effort required to design diverse scenarios.
While procedural map generation offers scalable diversity, no prior work systematically compares how different generator types affect policy generalization.
We integrate four generators (sparse, maze, graph, and \ac{WFC}) with guaranteed navigability into MuRoSim, a 2D simulator focusing on training efficiency for LiDAR-based navigation.
We cross-evaluate five navigation policies on 1000 seeded maps per generator across three training seeds.
Results show a strongly asymmetric cross-generator transfer: a specialist trained on sparse layouts falls to \SI{3.3}{\percent} success on mazes, whereas a policy trained on the combined generator set achieves \SI{91.5 \pm 1.1}{\percent} mean success.
We further demonstrate that $A^*$ path-planner subgoal inputs are the dominant factor for robustness, raising success from the \SI{90.2 \pm 1.4}{\percent} feedforward baseline to \SI{98.9 \pm 0.4}{\percent} and outperforming \ac{GRU} recurrence, which only improves the reactive baseline.
The \ac{DRL} policies outperform a classical Carrot+$A^*$ controller, which matches their success only at low speeds (\SI{1.0}{\meter\per\second}) but collapses to \SI{24.9}{\percent} at \SI{2.0}{\meter\per\second}. This highlights learned speed adaptation as the decisive advantage of the learned approach.
Real-world experiments on a RoboMaster confirm sim-to-real transfer in a cluttered arena, while a maze-like layout exposes remaining failure modes that recurrence helps mitigate.
\end{abstract}


\acresetall

\section{Introduction}

\begin{figure}[!t]
\centering
\includegraphics[width=\columnwidth]{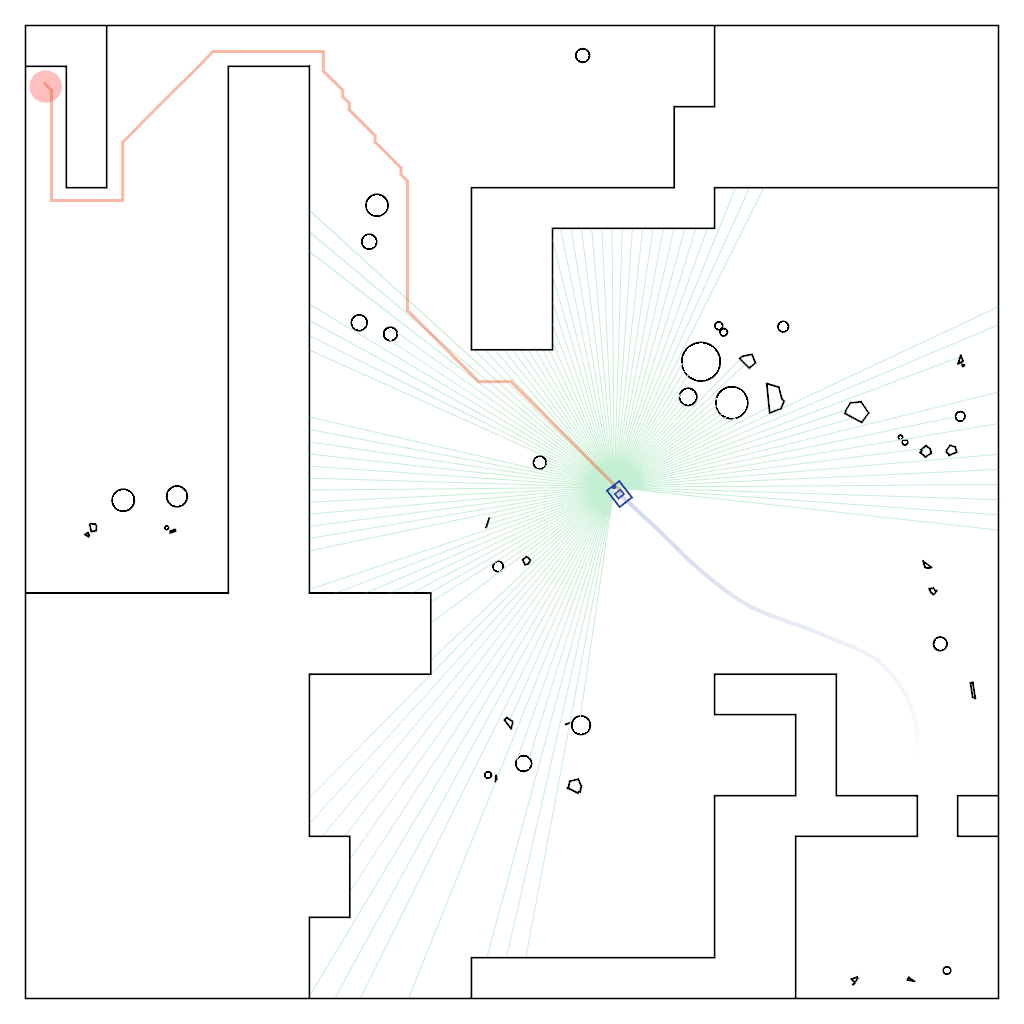}
\caption{A \ac{DRL} policy navigating a procedurally generated \ac{WFC} warehouse layout in MuRoSim. The robot follows the $A^*$ subgoal path (orange) toward the goal (red) while reacting to LiDAR observations (green); the trajectory tail (blue) shows the executed motion.}
\label{fig:teaser}
\end{figure}

In industrial and service robotics, mobile robots must navigate reliably across a wide range of environments, yet achieving robust generalization to structurally diverse settings remains a significant challenge.
Classical navigation stacks combine a global planner such as $A^*$ with a local planner such as the Dynamic Window Approach~\cite{foxDynamicWindowApproach1997}.
These methods rely on pre-built maps and hand-tuned parameters, making them brittle in environments that differ from the original configuration.
\ac{DRL} offers an alternative by learning reactive navigation policies directly from sensor data~\cite{taiVirtualtorealDeepReinforcement2017, jestelObtainingRobustControl2021}.

Despite this potential, \ac{DRL} policies are often evaluated on a narrow set of training environments, leading to poor generalization beyond the training distribution~\cite{kirkSurveyZeroshotGeneralisation2023}.
A primary reason for this limitation is the manual effort required to design diverse and structurally complex navigation scenarios.
One technique to address this issue is procedural generation of training environments~\cite{cobbeLeveragingProceduralGeneration2020}.
Procedural map generation provides a way to scale environment diversity at no manual cost.
By using algorithmic rules and different generator types, it is possible to produce fundamentally different obstacle layouts: open spaces with scattered objects, narrow maze corridors, irregular graph-based passages, or structured tile-based environments.
Despite the growing adoption of procedural generation for \ac{DRL} training, no prior work systematically evaluates how the choice of generator affects the resulting navigation policy's robustness.
For instance, it is unclear whether a policy trained on complex mazes inherently generalizes to open spaces, or if a combined training approach is necessary to achieve robust generalization.

Simply increasing the number and diversity of training environments is often not enough.
As the complexity of these generated layouts grows the limitations of purely end-to-end navigation policies become apparent.
While these reactive policies excel at local obstacle avoidance, they often struggle with the long-horizon planning required to navigate through intricate corridors or escape local minima, as we previously observed in dead-end sim-to-real scenarios~\cite{jestelMuRoSimFastEfficient2024}.
To overcome this, policies require additional guidance, such as a global planner providing subgoals or a memory mechanism to handle partial observability and long-term dependencies.
Which of these mechanisms contributes most to navigation performance across various layout types has not been systematically compared.

In this paper, we address these gaps with the following contributions:
\begin{itemize}
    \item We extend MuRoSim~\cite{jestelMuRoSimFastEfficient2024}, our custom lightweight 2D LiDAR navigation simulator, with a procedural map generator that produces navigability-guaranteed layouts spanning four structural classes (sparse, maze, graph, and \ac{WFC}); Fig.~\ref{fig:teaser} shows a rollout in a \ac{WFC} warehouse layout.
    \item Through a $5 \times 4$ cross-evaluation of policies trained on each generator and on a combined set, we show that cross-generator transfer is strongly asymmetric and that training on a combined set yields the most robust policy.
    \item We isolate the effect of $A^*$ subgoal direction inputs and \ac{GRU} recurrence on the combined policy, showing that subgoal inputs are the dominant factor for robust navigation.
    \item We show that the \ac{DRL} policies' main advantage over classical controllers is learned speed adaptation, enabling high-speed operation up to \SI{3.0}{\meter\per\second}.
    \item Real-world experiments on a RoboMaster transfer the trained policies without retraining: the combined policy navigates a cluttered arena reliably, while a maze-like layout exposes a dead-end failure mode that the recurrent policy resolves.
\end{itemize}

\section{Related Work}

\begin{figure*}[t]
    \centering
    \begin{subfigure}[t]{0.24\textwidth}
        \centering
        \includegraphics[width=\textwidth]{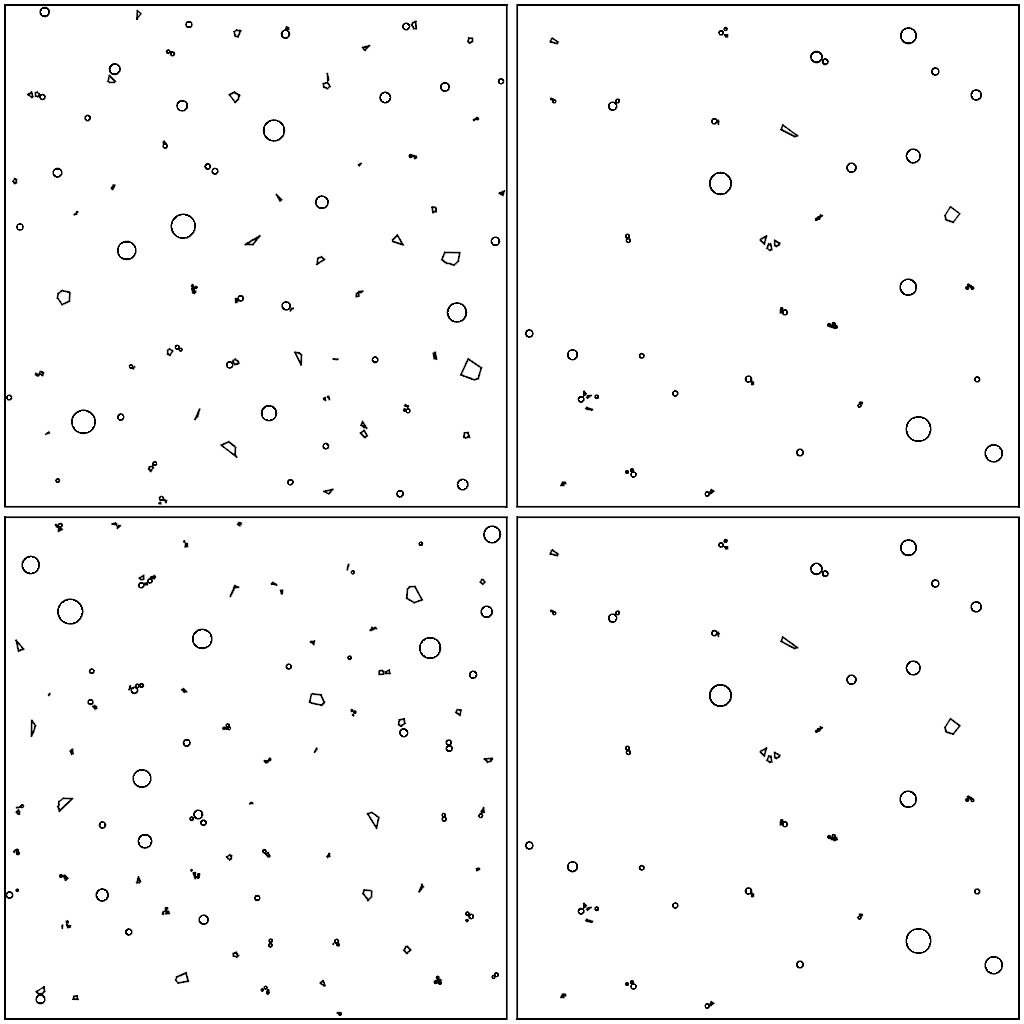}
        \caption{Sparse}
        \label{fig:sparse}
    \end{subfigure}\hfill
    \begin{subfigure}[t]{0.24\textwidth}
        \centering
        \includegraphics[width=\textwidth]{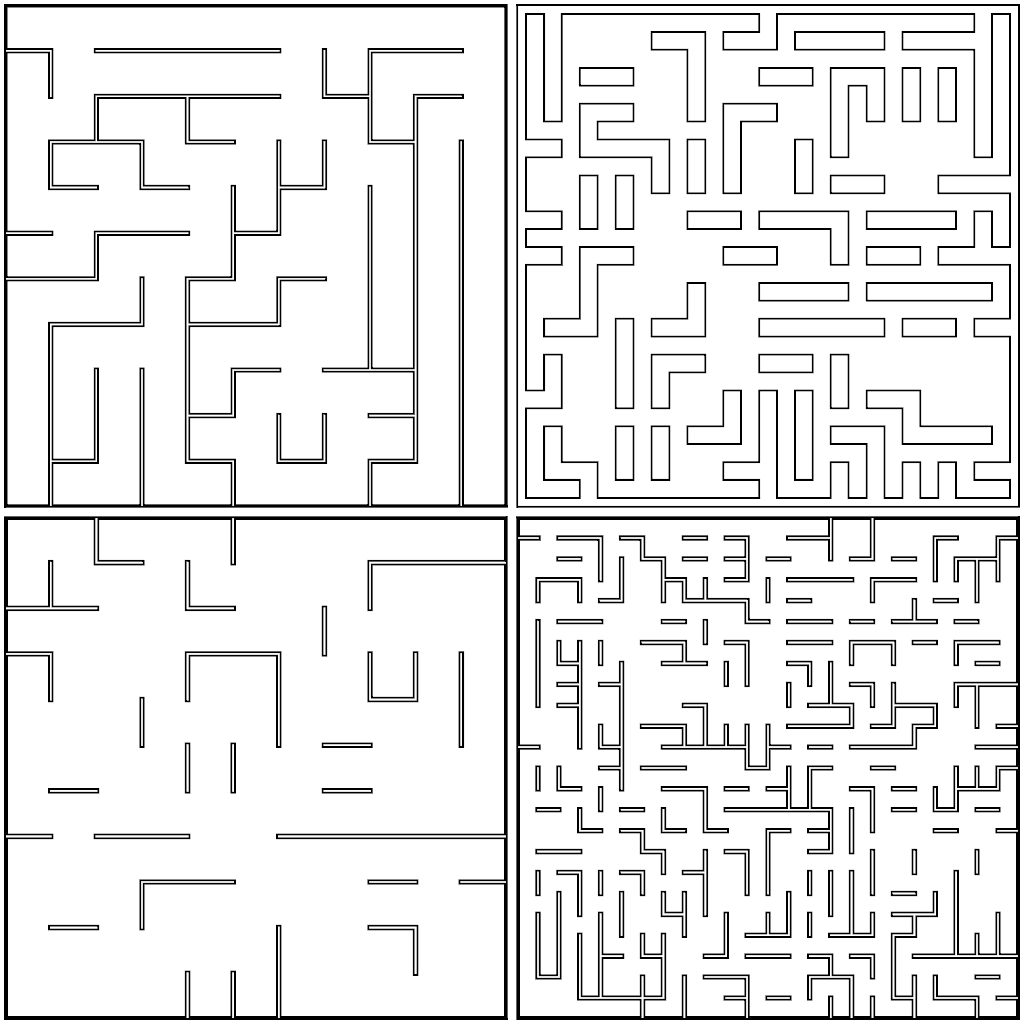}
        \caption{Maze}
        \label{fig:maze}
    \end{subfigure}\hfill
    \begin{subfigure}[t]{0.24\textwidth}
        \centering
        \includegraphics[width=\textwidth]{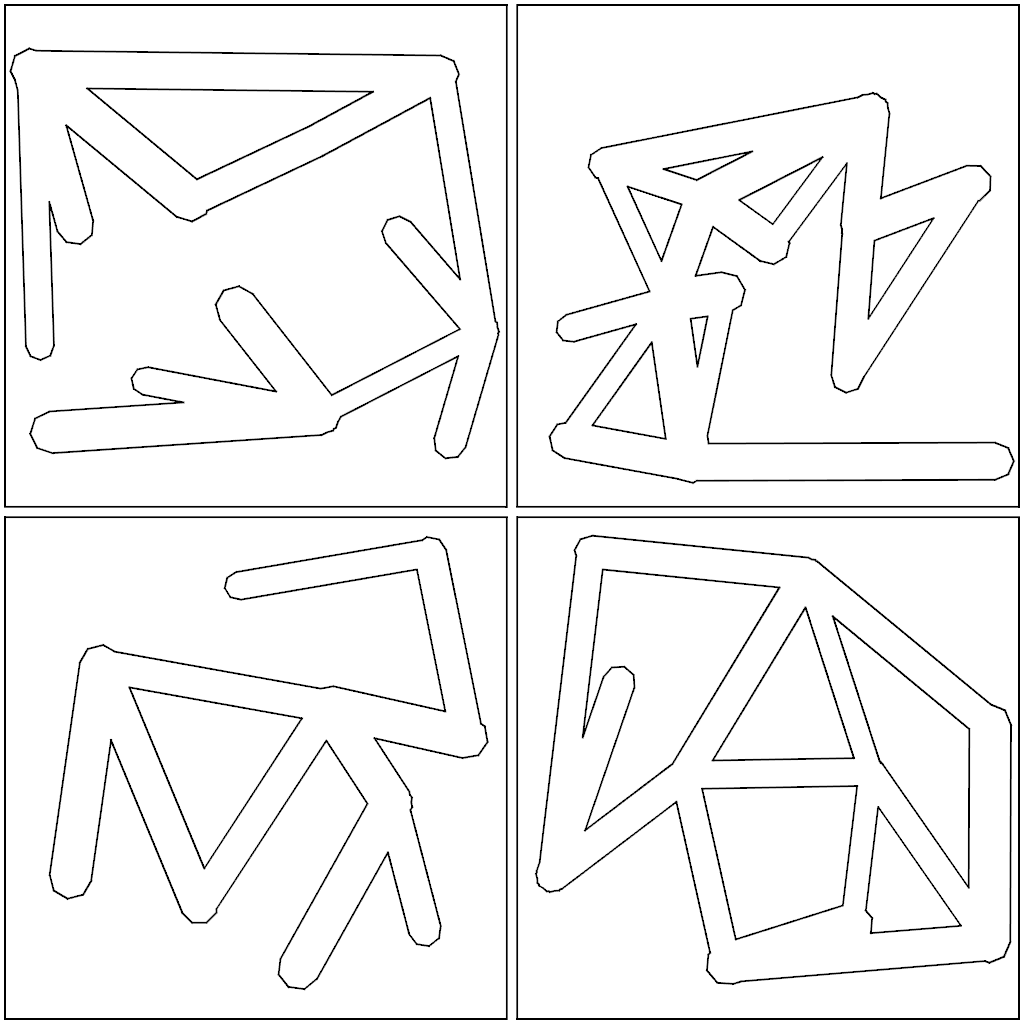}
        \caption{Graph}
        \label{fig:graph}
    \end{subfigure}\hfill
    \begin{subfigure}[t]{0.24\textwidth}
        \centering
        \includegraphics[width=\textwidth]{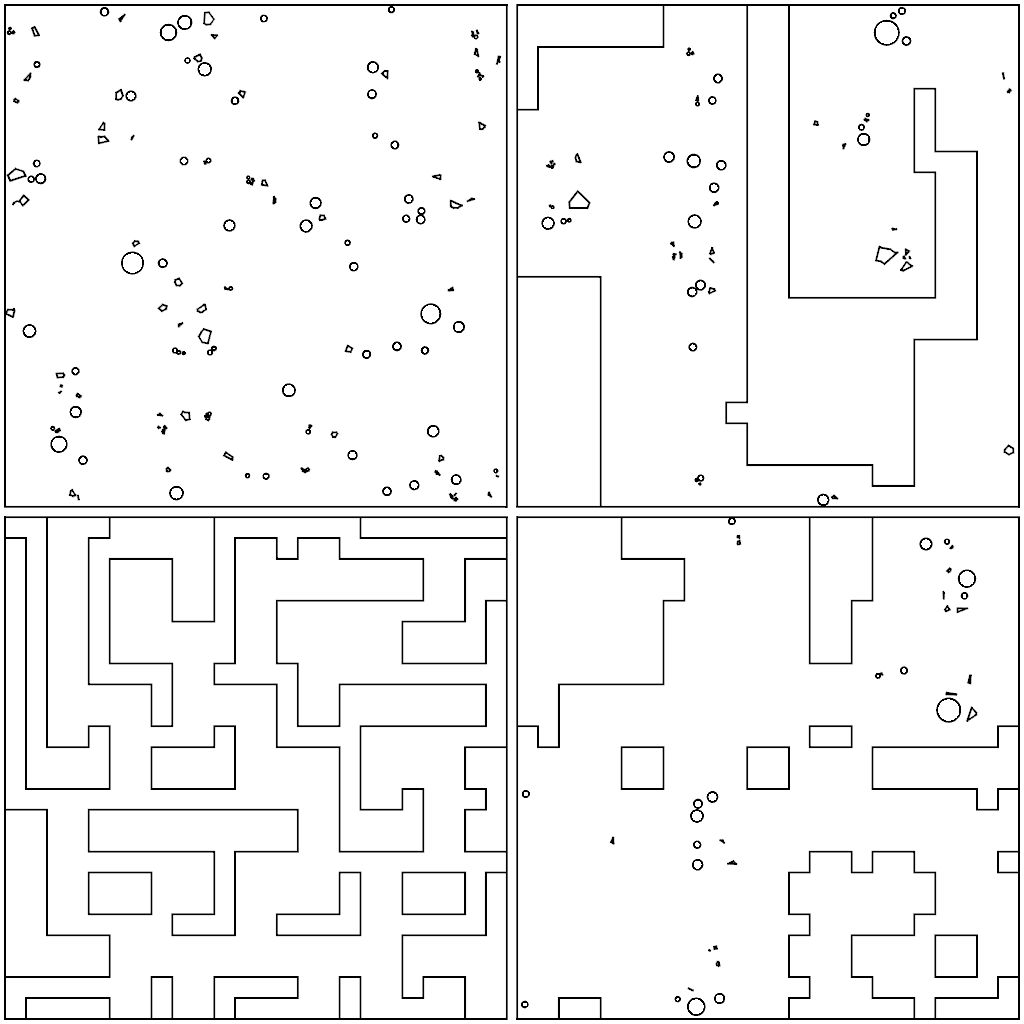}
        \caption{WFC (warehouse)}
        \label{fig:wfc}
    \end{subfigure}
    \caption{Example environments from the four procedural map generators. (a)~Sparse: randomly placed obstacles of varying shape and size. (b)~Maze: corridor-based layout generated via Prim's algorithm with wall removal. (c)~Graph: passage network from randomly connected seed points. (d)~WFC: tile-based layout using Wave Function Collapse with the four different presets.}
    \label{fig:generators}
\end{figure*}

\subsection{Deep Reinforcement Learning for Robot Navigation}

\ac{DRL} has become a widely used approach for training robots on a range of complex tasks, including navigation for mobile robots.
Early work demonstrated that mapless motion planners can be trained end-to-end in simulation and transferred to real robots using only sparse LiDAR readings~\cite{taiVirtualtorealDeepReinforcement2017}.
Subsequent studies have extended \ac{DRL} navigation to specialized settings: combining LiDAR data with velocity obstacles for dense pedestrian crowds~\cite{xieDRLVOLearningNavigate2023}, and scaling sensor-level collision avoidance to up to 100 robots in complex environments~\cite{fanDistributedMultirobotCollision2020}.
In contrast to our prior work on decentralized multi-robot policies~\cite{jestelObtainingRobustControl2021}, this paper isolates a single robot to avoid biasing the evaluation with inter-robot collisions and to focus on environment generalization.

\subsection{Simulator Tools for \ac{DRL}-based Navigation}
Simulation tools for \ac{DRL}-based navigation range from high-fidelity 3D platforms to lightweight 2D simulators.
High-fidelity simulators such as Gazebo provide full 3D physics and rendering but incur substantial computational overhead that limits training throughput~\cite{kaupReviewNinePhysics2024}, while Isaac Lab~\cite{nvidiaIsaacLabGPUAccelerated2025} addresses throughput with GPU acceleration but primarily targets manipulation and locomotion.
CAMAR~\cite{pshenitsynCAMARContinuousActions2025} is a standalone multi-agent reinforcement learning benchmark for continuous-action pathfinding.
MuRoSim~\cite{jestelMuRoSimFastEfficient2024} fills a niche by targeting CPU-only parallelism, exploiting the parallel instruction set \ac{SIMD}.
This allows thousands of environments for LiDAR-based navigation in parallel through multi-threaded stepping.
This efficiency is essential for evaluating generalization across the large-scale map sets used in our study.

\subsection{Procedural Content Generation}
Procedural content generation can produce the environment diversity needed for generalization for \ac{DRL} tasks~\cite{cobbeLeveragingProceduralGeneration2020}.
ProcTHOR~\cite{deitkeProcTHORLargeScale2022} demonstrated that training on 10,000 procedurally generated 3D houses produces state-of-the-art results across multiple embodied AI benchmarks, but targets visual tasks rather than 2D navigation.
Arena 5.0~\cite{kastnerDemonstratingArena502025} takes a different approach, using transformer-based models to generate social navigation scenarios from text prompts in 3D.
For 2D environments, Flores-Aquino et al.~\cite{flores-aquino2DGridMap2021} proposed dungeon-style grid map generation for navigation research.
Several algorithms target 2D layout generation: \ac{WFC}~\cite{karthWaveFunctionCollapseConstraintSolving2017} assembles locally consistent tile-based layouts from adjacency rules, Prim's algorithm~\cite{primShortestConnectionNetworks1957} produces corridor-based mazes with tunable complexity, and graph-based approaches such as Level Graph~\cite{vonrymonlipinskiLevelGraphIncremental2019} use minimum spanning trees to connect rooms and corridors.
While each of these methods has been used individually, the effect of the generator choice on the generalization of the resulting navigation policy has not been systematically studied.

\subsection{Hybrid Navigation Approaches}
To handle complex layouts, hybrid approaches often combine a global planner with a learned local policy to improve navigation in complex environments~\cite{debnathHybridApproachIndoor2025}.
Recurrent architectures such as \ac{GRU} are common baselines for partially observable \ac{DRL} tasks~\cite{niRecurrentModelFreeRL2022}.
Our work systematically compares global subgoal inputs versus temporal memory to determine which contributes most to robust navigation across structurally diverse layouts.
\section{Methodology}

This section details an experimental framework designed to evaluate policy robustness, integrating a highly efficient simulator, a suite of procedural map generators, and a hybrid network architecture.

\subsection{MuRoSim Simulator}

We utilize MuRoSim~\cite{jestelMuRoSimFastEfficient2024}, a lightweight C++ 2D multi-robot simulator with Python bindings to interface with Python-based \ac{DRL} frameworks.
By leveraging \ac{SIMD} instructions and multi-threaded simulation stepping and sensor computation, MuRoSim enables the CPU-based simultaneous execution of thousands of environments on a single machine.
To facilitate zero-shot sim-to-real transfer, the simulator uses continuous action and observation spaces with line-segment and circle obstacles, avoiding the artifacts inherent in grid-cell discretizations.
Additionally, MuRoSim discretizes the geometry into a grid-based occupancy map for utility functions such as $A^*$ path planning, randomized start/goal placement, and global observations for 2D-\ac{CNN} network architectures.
In our setup, robots are modeled as 2D holonomic or differential-drive agents equipped with configurable LiDAR sensors.
A central component of our methodology is the direct integration of procedural map generators into MuRoSim, as detailed in Section~\ref{sec:generators}, to facilitate the development of more generalizable \ac{DRL} navigation policies.

\subsection{Procedural Map Generators}
\label{sec:generators}

The structural characteristics of the training environment shape the behavior and generalization capabilities of the learned policy.
To investigate this relationship, we implement four procedural generators, each designed to synthesize a distinct class of obstacle layouts (Fig.~\ref{fig:generators}).
Several generator parameters are specified as ranges, from which values are sampled uniformly at random during generation to increase environment diversity.
All generators share the following common parameters:
\begin{itemize}
    \item \textit{node\_count}: Navigation nodes serving as start and goal positions, determining the maximum number of robots.
    \item \textit{node\_radius}: Free space around each node. Needs to be at least the robot radius to ensure collision-free placement.
    \item \textit{spacing}: Range parameter specifying the passable gap between corridors or obstacles. Must be large enough for the robot to navigate between any two nodes.
    \item \textit{world\_size}: Range parameter for the side length of the square environment, which can be constrained to fit within a fixed-size grid map observation.
\end{itemize}

\textbf{Sparse:}
The sparse generator (Fig.~\ref{fig:sparse}) places obstacles of random size and shape at random positions within the environment.
The spacing parameter guarantees a minimum distance between obstacles so the robot can navigate around them.
Obstacle shapes include circles, convex polygons, and decomposed polygons, with a density parameter controlling the number of obstacles.
This generator produces the simplest environments, representing open spaces with scattered obstacles commonly used in early \ac{DRL} navigation research~\cite{taiVirtualtorealDeepReinforcement2017}.

\textbf{Maze:}
The maze generator (Fig.~\ref{fig:maze}) creates corridor-based environments using Prim's algorithm~\cite{primShortestConnectionNetworks1957} to produce a perfect maze with no loops on a grid.
A wall removal rate parameter randomly removes a fraction of the walls, creating loops and alternative paths that reduce the number of dead ends.
The spacing parameter controls the corridor width.
A wall thickness parameter varies the thickness of the walls between environments, adding structural diversity.
Because a perfect maze contains exactly one path between any two cells, dead ends force the agent to fully backtrack to find an alternative route.
The wall removal rate mitigates this by introducing loops, but at low rates the dead-end-heavy structure makes mazes structurally demanding for reactive navigation policies.

\textbf{Graph:}
The graph generator (Fig.~\ref{fig:graph}) places random seed points and removes a fraction of them according to a point removal rate parameter.
The remaining points are connected with a Voronoi-like graph~\cite{aurenhammerVoronoiDiagramsSurvey1991}, and a minimum spanning tree ensures full connectivity.
Additional edges beyond the spanning tree are candidates for removal via an edge removal rate parameter.
The edges are extruded into corridors whose width is determined by the spacing parameter.
The resulting environments have irregular intersections at varying angles, producing passage layouts that are structurally between open sparse environments and constrained mazes.

\textbf{\ac{WFC}:}
The \ac{WFC} generator (Fig.~\ref{fig:wfc}) uses tile-based procedural generation~\cite{karthWaveFunctionCollapseConstraintSolving2017} with four presets: obstacle, labyrinth, warehouse, and cavern.
Each preset defines tile types and adjacency constraints that the \ac{WFC} solver uses to produce a globally consistent bitmap layout.
The bitmap distinguishes occupied cells (walls), free cells (navigable space), and buffer cells where random obstacles may be placed.
The largest connected free-space region is selected as the navigable area, since disconnected regions cannot be reached.
The \ac{WFC} generator produces the widest variety of environments, combining open areas, narrow corridors, and regions with randomly placed obstacles in a single layout.

\subsection{Robot Platform \& Network Architecture}
\label{sec:network}

We utilize a modified DJI RoboMaster S1 as our reference platform, following the configuration established in our previous work~\cite{jestelMuRoSimFastEfficient2024}.
The robot is a mecanum-wheeled platform capable of holonomic actuation, modeled with a rectangular footprint of $\SI{0.32}{\meter} \times \SI{0.24}{\meter}$ in the simulator.
It is equipped with an RPLidar S2 2D LiDAR sensor providing a \SI{360}{\degree} field of view with 3200 laser rays and a maximum sensing range of \SI{30}{\meter}.
To isolate the impact of environmental structure on policy performance and eliminate confounds from inter-robot interactions, each training and evaluation scenario is limited to a single agent.

\textbf{Observation:} The observation consists of a LiDAR scan cropped to a \SI{270}{\degree} field of view with $1200$ rays (normalized) and scalar values including the unit-vector orientation to the goal, distance to the goal, linear velocity, and the previous commanded velocity.
For experiments with $A^*$ subgoal inputs, the observation is extended with five unit vectors describing the next five \SI{0.5}{\meter} segments of the $A^*$ path: the first vector points from the robot to the first waypoint, and each subsequent vector points from one waypoint to the next.

\textbf{Action:} The continuous action space consists of commanded linear velocities $(v_x, v_y)$ and angular velocity $\omega$, with $v_x \in [-0.5, 3.0]\,\si{\meter\per\second}$, $v_y \in [-1.0, 1.0]\,\si{\meter\per\second}$, and $\omega \in [-2.0, 2.0]\,\si{\radian\per\second}$.
The forward range is larger than the reverse range because the \SI{270}{\degree} LiDAR field of view leaves a \SI{90}{\degree} blind spot behind the robot, so high-speed reverse motion would be unsafe.

\textbf{Network:} We use \ac{PPO}~\cite{schulmanProximalPolicyOptimization2017} with separate actor and critic networks.
Both networks share the same architecture, consisting of two separate input streams for the LiDAR and scalar observations.
The LiDAR stream is processed by three 1D convolutional layers inspired by the IMPALA \acs{CNN} architecture~\cite{espeholtIMPALAScalableDistributed2018}, followed by a linear layer producing a 256-dimensional feature vector.
The three layers use $32$, $24$, and $16$ filters with kernel sizes $9$, $7$, $5$ and strides $4$, $3$, $2$ respectively, all with ReLU activations.
The scalar stream is concatenated into one vector and processed by a separate \ac{MLP} producing a 96-dimensional vector.
Both streams are concatenated and passed through a fusion \ac{MLP} that outputs a 256-dimensional latent representation.
The actor output layer maps this to the action mean $\boldsymbol{\mu} \in \mathbb{R}^3$ with a state-independent learnable log-standard deviation, while the critic outputs a single scalar value estimate.
For the baseline network, we use frame stacking of 3 frames for both the LiDAR and scalar observations.
For the recurrent network variant without frame stacking, we use a \ac{GRU} layer with a hidden size of 256 after the fusion \ac{MLP}.

\textbf{Reward Function:} The reward at step $t$ has the form
\begin{equation}
    r_t = r^{\Delta d}_t + r^{\text{laser}}_t + r^{\text{goal}}_t + r^{\text{action}}_t + r^{\text{done}}_t,
    \label{eq:reward}
\end{equation}
following the structure of our prior work~\cite{jestelObtainingRobustControl2021} with weights re-tuned via an Optuna~\cite{akibaOptunaNextgenerationHyperparameter2019} hyperparameter sweep.
The progress term $r^{\Delta d}_t$ rewards a decrease in a reference distance and penalizes an increase: in Experiment~1 the reference is the straight-line distance to the goal, in Experiment~2 it is the distance to a \SI{0.5}{\meter} lookahead point along the $A^*$ path.
The laser term $r^{\text{laser}}_t$ applies a small linear penalty whenever the minimum over all LiDAR rays falls below a safety threshold.
The goal term $r^{\text{goal}}_t$ applies a continuous bonus while the robot is inside the goal radius, growing linearly as the robot approaches the goal center.
The action-rate term $r^{\text{action}}_t$ penalizes large changes between consecutive commanded velocities to promote smooth motion.
The done term $r^{\text{done}}_t$ ends the episode and applies a positive reward on goal reach, a negative reward on collision, and zero on timeout.

\section{Experiments}

\begin{table*}[t]
\caption{Cross-evaluation success, collision, and timeout rates (\%). Each cell is the 3-seed mean over 1000 evaluation maps; the Mean column adds the std across seeds.}
\centering
\resizebox{\textwidth}{!}{
\begin{tabular}{l@{\hskip 10pt}ccccc@{\hskip 14pt}ccccc@{\hskip 14pt}ccccc}
\toprule
 & \multicolumn{5}{c}{\textbf{Success Rate}} & \multicolumn{5}{c}{\textbf{Collision Rate}} & \multicolumn{5}{c}{\textbf{Timeout Rate}} \\
\cmidrule(lr){2-6} \cmidrule(lr){7-11} \cmidrule(lr){12-16}
\textbf{Policy} & \textbf{Sparse} & \textbf{Maze} & \textbf{Graph} & \textbf{WFC} & \textbf{Mean} & \textbf{Sparse} & \textbf{Maze} & \textbf{Graph} & \textbf{WFC} & \textbf{Mean} & \textbf{Sparse} & \textbf{Maze} & \textbf{Graph} & \textbf{WFC} & \textbf{Mean} \\
\midrule
Sparse   & \textbf{99.5} &  3.3 & 49.6 & 50.5 & $50.7 \pm 1.4$ &  0.5 & 96.7 & 50.3 & 49.3 & $49.2 \pm 1.3$ &  0.0 &  0.0 &  0.1 &  0.2 & $0.1 \pm 0.1$ \\
Maze     & 84.6 & 75.1 & 65.5 & 41.9 & $66.8 \pm 3.5$ &  9.4 &  2.8 &  5.0 & 17.4 & $8.7 \pm 3.2$ &  6.0 & 22.1 & 29.5 & 40.7 & $24.6 \pm 3.4$ \\
Graph    & 93.6 & 50.9 & 96.9 & 80.4 & $80.5 \pm 3.3$ &  6.4 & 29.0 &  0.7 & 12.6 & $12.2 \pm 3.6$ &  0.0 & 20.1 &  2.5 &  6.9 & $7.4 \pm 1.6$ \\
WFC      & 98.7 & 58.9 & 93.5 & \textbf{93.4} & $86.1 \pm 2.2$ &  1.3 & 18.9 &  3.9 &  1.7 & $6.4 \pm 1.4$ &  0.0 & 22.3 &  2.6 &  4.9 & $7.4 \pm 2.3$ \\
Combined & 99.1 & \textbf{76.7} & \textbf{97.0} & 93.2 & $\mathbf{91.5 \pm 1.1}$ &  0.9 &  4.2 &  1.0 &  2.9 & $2.2 \pm 0.7$ &  0.0 & 19.1 &  2.0 &  3.9 & $6.2 \pm 0.4$ \\
\bottomrule
\end{tabular}
}
\label{tab:cross_eval}
\end{table*}

\begin{figure*}[t]
    \centering
    \includegraphics[width=\textwidth]{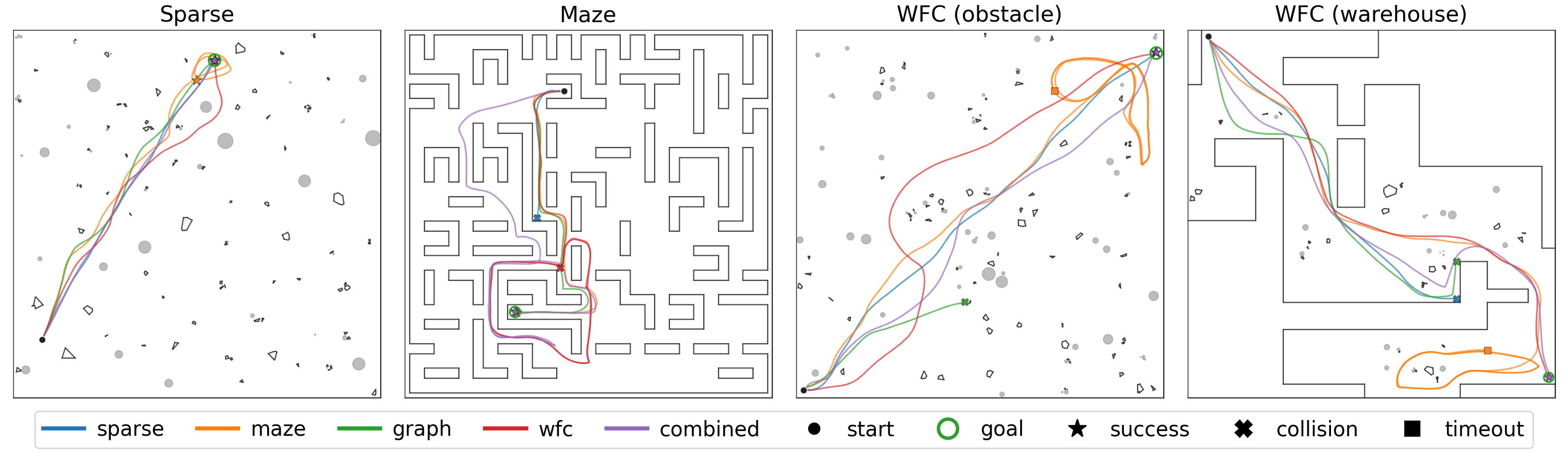}
    \caption{Trajectories of the five cross-evaluation policies on one representative map per environment type (left to right: sparse, maze, \acs{WFC} obstacle, \acs{WFC} warehouse). Colored lines show policy trajectories; end markers indicate outcome ($\star$ success, $\times$ collision, $\blacksquare$ timeout); the green ring marks the goal radius. The combined policy succeeds on all four maps, while each specialist fails on at least one out-of-distribution layout: the sparse specialist collides in cluttered maze and warehouse environments, the maze specialist's corridor-tuned speed produces visible orbital loops around the goal, and the graph specialist collides in both \acs{WFC} layouts.}
    \label{fig:cross_eval_traj}
\end{figure*}

In this section, we present the experimental setup and results for the two main experiments:
(1) a cross-evaluation of baseline navigation policies trained on different procedural generators to assess generalization, and
(2) an ablation study comparing the effect of $A^*$ subgoal direction inputs and a \ac{GRU} memory layer on navigation performance.

\subsection{Training and Evaluation Setup}

For the first experiment, we train five policies to evaluate the effect of the generator type on policy generalization.
All policies share the network architecture described in Section~\ref{sec:network}.
Four specialist policies are each trained exclusively on one generator (sparse, maze, graph, or \ac{WFC}).
A fifth combined policy is trained on all four generators simultaneously, with each generator sampled with equal probability.

Each configuration is trained with three independent random seeds, and all reported metrics are means across seeds with the mean-column standard deviation included to convey seed sensitivity.
Each seed is trained for $10^8$ environment steps using $4096$ parallel simulations with \ac{PPO} and \ac{GAE} ($\gamma = 0.99$, $\lambda_{\mathrm{GAE}} = 0.95$, clip ratio $0.2$, gradient-norm clip $0.5$, value loss coefficient $0.5$, one epoch per update).
The learning rate is fixed at $8 \times 10^{-4}$ and was tuned together with the reward scale factors via an Optuna~\cite{akibaOptunaNextgenerationHyperparameter2019} hyperparameter sweep.
We use a rollout horizon of $64$ steps and a mini-batch size of $8192$.
Episodes run for a maximum of $512$ steps at a simulation time step of \SI{0.1}{\second}, giving a maximum episode duration of \SI{51.2}{\second}.
Environments are regenerated with new random seeds every $1$M steps to maintain diversity throughout training, and evaluation is performed at the same interval on a separate set of seeded environments.
For the combined policy, one of the four generator families is sampled with equal probability per episode, with the \ac{WFC} family split evenly across its four presets (obstacle, labyrinth, warehouse, cavern).

All generators share a minimum node count of $5$ and a node radius of \SI{0.25}{\meter}, ensuring sufficient clearance at start and goal positions.
The world size is sampled from $[10.0, 20.0]\,\si{\meter}$ and the spacing from $[0.5, 1.25]\,\si{\meter}$ per environment.
For the graph and \ac{WFC} generators, we adjust the spacing range to $[0.7, 1.25]\,\si{\meter}$ and $[0.55, 1.0]\,\si{\meter}$, respectively, to ensure enough free cells in the grid map for $A^*$ planning in the second experiment.

For evaluation, each policy is run in deterministic mode (no exploration noise) on 1000 maps per generator generated with a fixed random seed for reproducibility across all configurations.
The same protocol is used for both experiments, with success, collision, and timeout rates recorded per episode.

\subsection{Cross-Evaluation Results}

Success, collision, and timeout rates for each policy-generator combination are reported in Table~\ref{tab:cross_eval}.
Each row represents a trained policy, each column the evaluation generator.
The diagonal entries correspond to in-distribution evaluation, where the policy is tested on its own training generator.

Four of the five policies achieve above \SI{93}{\percent} success on sparse environments; the maze specialist is the sole exception at \SI{84.6}{\percent}, a residue of the overshooting behavior that the corridor-trained policy learned to rely on for fast progress.
On each individual generator, the combined policy matches or narrowly exceeds the corresponding specialist on three of four columns: it narrowly trails the sparse specialist on sparse (\SI{99.1}{\percent} vs.\ \SI{99.5}{\percent}), overtakes the maze specialist on maze (\SI{76.7}{\percent} vs.\ \SI{75.1}{\percent}), and overtakes the graph specialist on graph (\SI{97.0}{\percent} vs.\ \SI{96.9}{\percent}); the \ac{WFC} specialist retains a narrow lead on \ac{WFC} (\SI{93.4}{\percent} vs.\ \SI{93.2}{\percent}).

Maze environments are the hardest across all policies.
The maze specialist reaches \SI{75.1}{\percent}, while the sparse and graph specialists drop to \SI{3.3}{\percent} and \SI{50.9}{\percent} respectively, and the \ac{WFC} specialist falls to \SI{58.9}{\percent}.
This confirms that narrow corridors require navigation skills that do not transfer from open or semi-structured environments.

The combined policy achieves the highest mean success rate of \SI{91.5 \pm 1.1}{\percent}, outperforming every specialist on average and exhibiting the smallest seed-to-seed variance of any policy.
It matches or narrowly trails the specialists on sparse and \ac{WFC} and exceeds them on maze and graph, while remaining robust on all four evaluation generators ($\geq \SI{76}{\percent}$).
Training on diverse generators therefore yields a policy that generalizes broadly without sacrificing in-distribution performance on the hardest generator type.

Failure modes differ strongly by specialist and generator type, as shown by the collision and timeout columns of Table~\ref{tab:cross_eval} and illustrated on representative maps in Fig.~\ref{fig:cross_eval_traj}.
The sparse specialist almost never times out (\SI{0.1}{\percent} mean) but fails predominantly through collision off-distribution, peaking at \SI{96.7}{\percent} on maze environments where the policy drives straight toward the goal without evading walls.
The maze specialist exhibits a different pattern with a \SI{24.6}{\percent} mean timeout rate and \SI{8.7}{\percent} mean collision rate: its corridor-tuned speed causes it to overshoot the goal when no walls are nearby, after which the agent orbits the goal until the 512-step budget expires.
The combined policy keeps collision rates below \SI{5}{\percent} on every generator and incurs its largest timeout rate of \SI{19.1}{\percent} on maze environments, where precise long-horizon maneuvering remains difficult.

On sparse and graph environments, the few failures that occur are predominantly collisions rather than timeouts.
Out-of-distribution specialists show sharply higher collision rates than the combined policy, most strikingly the sparse specialist on maze (\SI{96.7}{\percent} collision), which was never trained in walled environments, and the graph specialist on maze (\SI{29.0}{\percent} collision), whose irregular-angle corridors do not prepare it for the tight right-angle passages of mazes.

\subsection{Network Architecture and Planner Comparison}

As in the first experiment, we use the baseline navigation policy architecture described in Section~\ref{sec:network}, which supports reactive navigation but lacks explicit planning, resulting in high timeout rates on maze environments.
To evaluate the effect of providing global path information, we extend the baseline with subgoal direction inputs derived from an $A^*$ planner.
We also evaluate adding a \ac{GRU} memory layer, which can provide temporal context and help with partial observability.
Four \ac{DRL} configurations are trained on the combined generator set for $100$M steps each: a feedforward baseline (\ac{FF}), a \ac{GRU} baseline, a feedforward policy with subgoal input (\acs{FF}+\ac{SG}), and a \acs{GRU} policy with subgoal input (\acs{GRU}+\acs{SG}).
Each subgoal configuration receives a 10-dimensional vector encoding the directions to five waypoints along the $A^*$ planned path at \SI{0.5}{\meter} intervals.
All four configurations share the path-aware progress reward $\Delta d_{\mathrm{path}}$ that credits motion along the $A^*$ path rather than straight-line distance to the goal, so the \acs{FF} / \acs{GRU} comparison isolates the effect of $A^*$ as an \emph{observation} input rather than as a reward signal.

The \ac{FF} configurations reuse the training hyperparameters of the first experiment (rollout horizon $64$, mini-batch $8192$, $4096$ parallel simulations, LiDAR frame stack of $3$).
The \ac{GRU} configurations instead use a rollout horizon of $256$ steps, a mini-batch of $512$, a backpropagation-through-time sequence length of $64$, and $2048$ parallel simulations; the LiDAR frame stack is reduced to $1$ since temporal context is provided by the recurrent layer.

As a classical baseline, we implement a simple Carrot pure-pursuit controller following the same $A^*$ path: a point on the path at a fixed lookahead distance of \SI{0.5}{\meter} is treated as the tracking target, a proportional heading controller steers toward it, and forward speed is held at a fixed cap.
We report the Carrot controller at three top-speed caps to isolate the effect of speed adaptation: a safe \SI{1.0}{\meter\per\second} at which the controller operates without timeouts, and two higher caps of \SI{1.5}{\meter\per\second} and \SI{2.0}{\meter\per\second}, all well below the \ac{DRL} policies' \SI{3.0}{\meter\per\second} cap.
The controller uses an $A^*$ planning grid with \SI{0.05}{\meter} resolution and a \SI{0.20}{\meter} obstacle-inflation buffer sized to the robot footprint.

\begin{figure}[t]
    \centering
    \includegraphics[width=\linewidth]{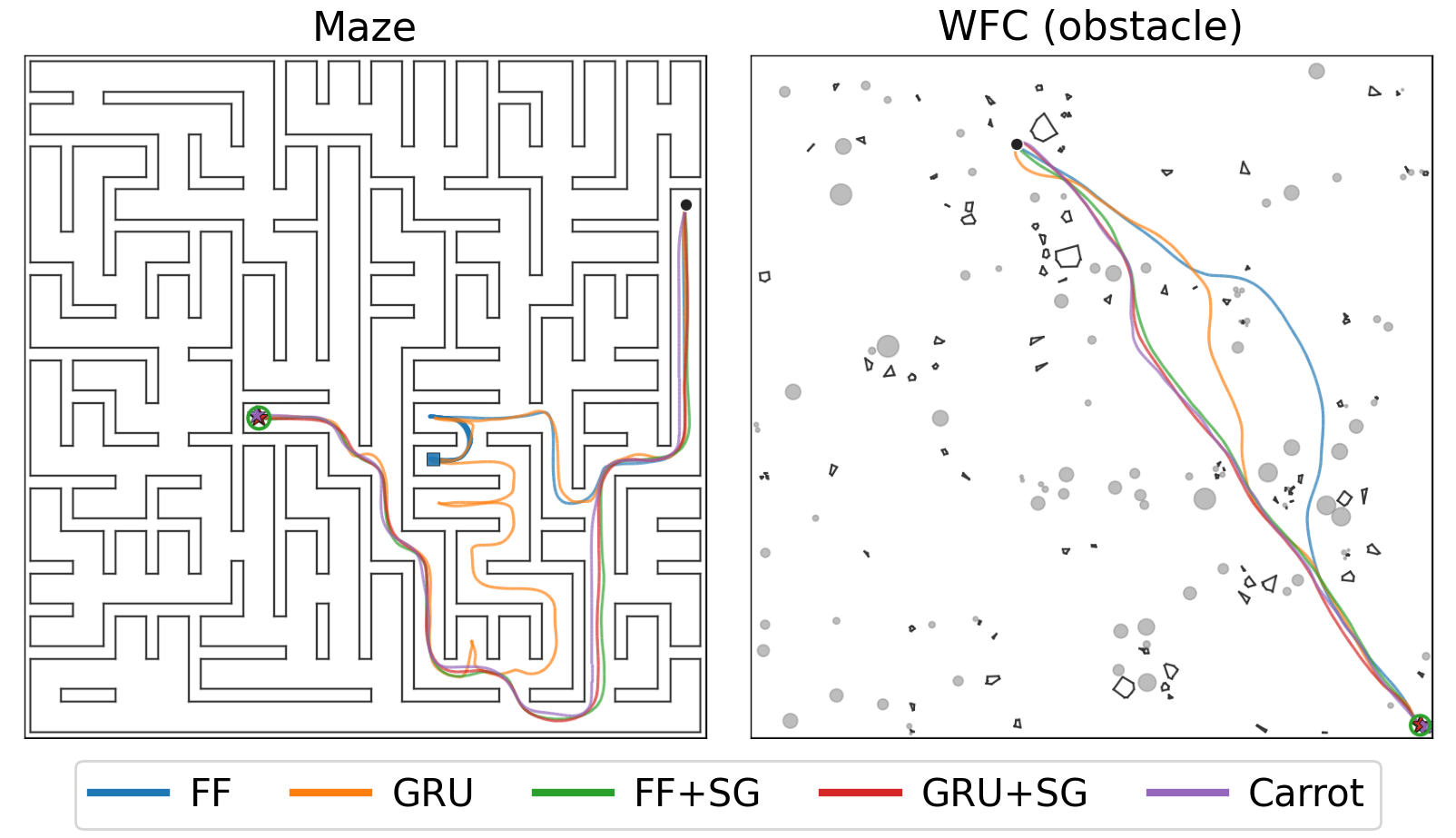}
    \caption{Trajectories of the four \acs{DRL} configurations and the Carrot controller (at \SI{1.0}{\meter\per\second}) on a maze (left) and a \acs{WFC} obstacle map (right); markers as in Fig.~\ref{fig:cross_eval_traj}. The baselines struggle on the maze: \acs{FF} (blue) times out and \acs{GRU} (orange) only escapes after extensive exploration, while the subgoal policies and Carrot track the $A^*$ path cleanly. On the \acs{WFC} map, the subgoal policies and Carrot follow a near-optimal path, but Carrot collides at the goal because its fixed-speed tracker cannot decelerate before the final obstacle.}
    \label{fig:network_eval_traj}
\end{figure}

\begin{table}[htbp]
\caption{Network architecture comparison: success rates (\%). Each \ac{DRL} cell is the 3-seed mean over 1000 maps; the Mean column adds the std across seeds. Carrot is deterministic (single run).}
\begin{center}
\begin{tabular}{l@{\hskip 6pt}ccccc}
\toprule
\textbf{Config} & \textbf{Sparse} & \textbf{Maze} & \textbf{Graph} & \textbf{WFC} & \textbf{Mean} \\
\midrule
Carrot (\SI{1.0}{\meter\per\second}) & 99.6 & 96.8 & 98.4 & 92.5 & 96.8 \\
Carrot (\SI{1.5}{\meter\per\second}) & 97.6 & 34.0 & 68.1 & 46.8 & 61.7 \\
Carrot (\SI{2.0}{\meter\per\second}) & 83.4 &  0.8 & 12.5 &  2.8 & 24.9 \\
\midrule
FF               & \textbf{99.7} & 70.3 & 97.4 & 93.3 & $90.2 \pm 1.4$ \\
GRU              & 98.7 & 81.0 & 98.1 & 96.0 & $93.5 \pm 0.8$ \\
FF + Subgoal     & 98.9 & \textbf{98.8} & \textbf{99.9} & \textbf{97.9} & $\mathbf{98.9 \pm 0.4}$ \\
GRU + Subgoal    & 98.9 & 97.0 & 99.6 & 97.6 & $98.3 \pm 0.5$ \\
\bottomrule
\end{tabular}
\label{tab:network_eval}
\end{center}
\end{table}

As shown in Table~\ref{tab:network_eval}, $A^*$ subgoal direction inputs are the dominant factor for improving navigation performance, and the resulting trajectory differences are illustrated on representative maps in Fig.~\ref{fig:network_eval_traj}.
The feedforward subgoal policy (\acs{FF}+\acs{SG}) achieves \SI{98.9 \pm 0.4}{\percent} mean success, compared to \SI{90.2 \pm 1.4}{\percent} for the baseline without subgoal input.
The largest improvement occurs on maze environments, where \acs{FF}+\acs{SG} reaches \SI{98.8}{\percent} compared to \SI{70.3}{\percent} for the baseline.
Subgoal variants also exhibit the smallest seed-to-seed variance ($\leq \SI{0.5}{\pp}$), showing that $A^*$ path information stabilizes training and reduces sensitivity to the initialization seed.

The \ac{GRU} memory layer provides a clear improvement to the baseline (\SI{93.5}{\percent} vs \SI{90.2}{\percent}), concentrated on maze environments where temporal context helps recover from dead ends (\SI{81.0}{\percent} vs \SI{70.3}{\percent}), but offers no additional benefit on top of the subgoal variant (\SI{98.3}{\percent} vs \SI{98.9}{\percent}).

The classical Carrot controller approaches the \ac{DRL} policies only at \SI{1.0}{\meter\per\second}, reaching \SI{96.8}{\percent} mean success, still \SI{2.1}{\pp} below the \acs{FF}+\acs{SG} policy and at one-third of the \ac{DRL} policies' \SI{3.0}{\meter\per\second} cap.
Success drops sharply as the cap is raised: \SI{61.7}{\percent} mean at \SI{1.5}{\meter\per\second} and \SI{24.9}{\percent} mean at \SI{2.0}{\meter\per\second}, where the controller collides on more than \SI{85}{\percent} of non-sparse maps because it cannot slow down before tight passages or sharp turns.
On successful episodes, the subgoal policies reach the goal in \SIrange{7.7}{7.8}{\second} on average with a traveled-to-$A^*$-path ratio of \numrange{1.07}{1.08}, while Carrot takes \SI{15.4}{\second} at a slightly tighter $1.02$ ratio; the baselines without subgoal input deviate further from the $A^*$ path (\numrange{1.23}{1.28} ratio) due to exploratory detours.

\subsection{Sim-to-Real Transfer}
To check that the trained policies are not specific to the simulator, we deploy several of them on a physical RoboMaster robot without retraining or fine-tuning.
Inference runs on a Jetson Orin NX at the same \SI{10}{\hertz} control rate used in training.
We first build a map of the lab with \texttt{slam\_toolbox}~\cite{macenskiSLAMToolboxSLAM2021} and then use Adaptive Monte Carlo Localization~\cite{foxAdaptingSampleSize2003} on the saved map, with an RPLidar~S2 as the only exteroceptive sensor.
The policy runs as a ROS~2~\cite{macenskiRobotOperatingSystem2022} action server, and the operator sends goal poses through RViz.

\begin{figure}[t]
    \centering
    \begin{subfigure}[t]{0.48\linewidth}
        \centering
        \includegraphics[width=\linewidth]{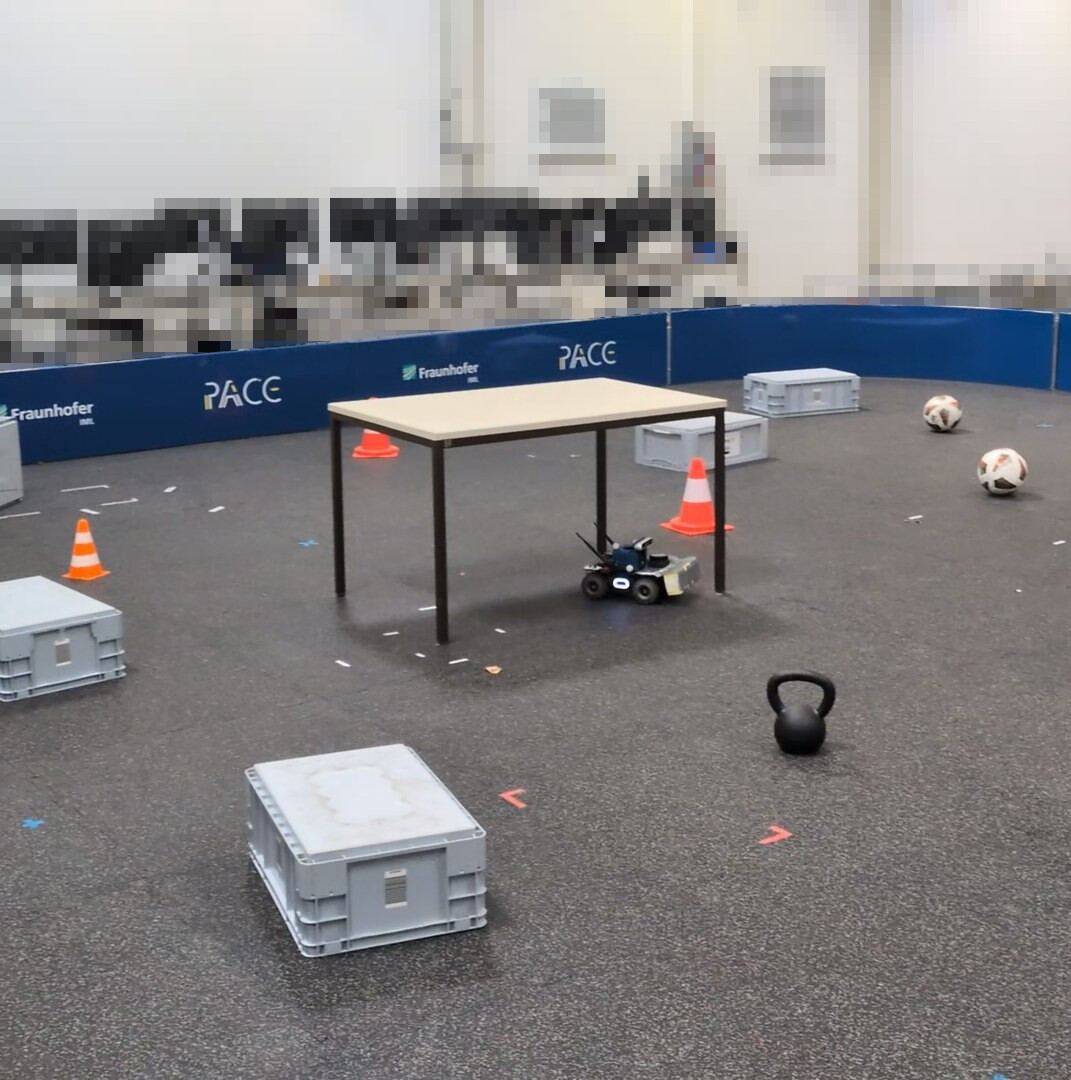}
        \caption{Cluttered open arena.}
        \label{fig:sim_to_real_arena}
    \end{subfigure}
    \hfill
    \begin{subfigure}[t]{0.48\linewidth}
        \centering
        \includegraphics[width=\linewidth]{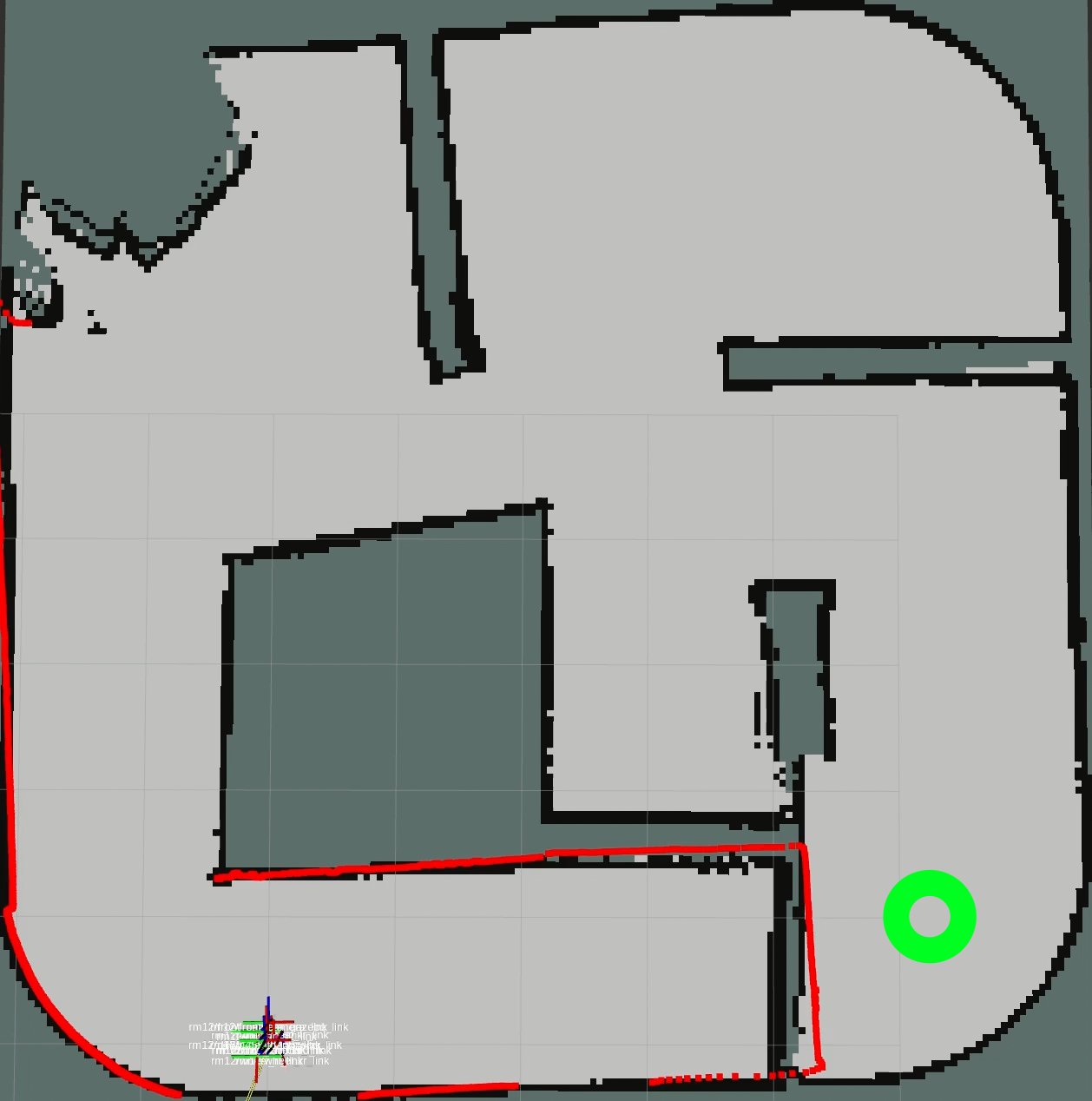}
        \caption{Maze-like environment in RViz.}
        \label{fig:sim_to_real_rviz}
    \end{subfigure}
    \caption{Sim-to-real runs on the RoboMaster.\protect\footnotemark{} (a)~Third-person view of the cluttered open arena used for the combined policy. (b)~RViz snapshot of the maze-like environment used to compare the sparse, combined, and \acs{GRU} policies, showing the \texttt{slam\_toolbox} map (gray), the live RPLidar scan (red), and the operator-set goal pose (green).}
    \label{fig:sim_to_real}
\end{figure}

We use two physical setups (Fig.~\ref{fig:sim_to_real}).
A successful run of the combined policy in a cluttered open arena containing tables, traffic cones, logistics containers, and footballs is shown in Fig.~\ref{fig:sim_to_real_arena}.
The policy reaches the operator-set goal while avoiding obstacle types that were never seen during training, and it deliberately passes close to obstacles to keep the path short.
In a second setup we build a maze-like environment (Fig.~\ref{fig:sim_to_real_rviz}) and compare three policies on it.
The sparse specialist drives straight at the goal and collides with the dividing wall.
The combined policy enters a dead end, returns toward the goal, drives into the same dead end a second time, and fails to reach the goal.
The \ac{GRU} policy enters the first dead end once, briefly explores the middle dead end, and then drives directly to the goal.
These runs extend the sim-to-real validation of our prior work~\cite{jestelMuRoSimFastEfficient2024} and remove its reliance on an external tracking system for localization.

\footnotetext{Video of the sim-to-real runs: \url{https://youtu.be/Rz1LFFEbgKg}}

\section{Discussion}

Generator diversity during training is the single most important factor for generalization.
The combined policy achieves the highest mean success rate and outperforms the corresponding specialist on graph and maze, while remaining within \SI{0.4}{\pp} of the specialists on sparse and \ac{WFC}.
Even with diverse training, mazes remain the hardest environment for every policy, exposing a fundamental limitation of reactive navigation: narrow corridors and dead ends require multi-step planning that a feedforward policy with only LiDAR and a few scalar inputs cannot perform.
The combined policy's high timeout rate on mazes (\SI{19.1}{\percent}) confirms it has learned to avoid collisions but cannot commit to long detours when the direct path is blocked, motivating the path-aware reward and observation modifications evaluated in the second experiment.

Providing $A^*$ subgoal directions as direct observation inputs is the single most effective modification to the policy.
The feedforward subgoal variant raises mean success from \SI{90.2}{\percent} to \SI{98.9}{\percent} and nearly eliminates the maze gap (\SI{70.3}{\percent} to \SI{98.8}{\percent}), with the smallest cross-seed variance in the experiment (\SIrange{0.4}{0.5}{\pp}).
A \ac{GRU} memory layer is a partial substitute: it improves the reactive baseline by \SI{3.3}{\pp} on average, concentrated on mazes, but provides no additional benefit once subgoal directions are already observed, suggesting that explicit planning input replaces the spatial reasoning that recurrence would otherwise need to learn.

The classical Carrot controller shares the same $A^*$ planner as the \ac{DRL} subgoal policies, so the gap between them isolates the contribution of the learned local behavior.
At a safe \SI{1.0}{\meter\per\second} top speed, Carrot reaches \SI{96.8}{\percent} mean success, just \SI{2.1}{\pp} below the feedforward subgoal policy; raising the cap to \SI{2.0}{\meter\per\second} collapses success to \SI{24.9}{\percent}, while the \ac{DRL} policies continue to operate reliably at their full \SI{3.0}{\meter\per\second} cap because they decelerate before tight passages.
Notably, the feedforward baseline without subgoal input (\SI{90.2}{\percent}) is beaten by the slow Carrot controller (\SI{96.8}{\percent}), confirming that the $A^*$ subgoal observation rather than the neural network alone is what lets \ac{DRL} outperform the classical baseline.

\section{Conclusion}

We presented a systematic evaluation of how procedural map generators and network architecture choices affect the generalization of \ac{DRL}-based robot navigation policies.

Across 1000 seeded maps per generator and three training seeds, specialist policies degraded sharply on structurally different layouts, while the combined policy reached \SI{91.5 \pm 1.1}{\percent} mean success with the smallest seed variance.
Adding $A^*$ subgoal directions as observation inputs raised success to at least \SI{97}{\percent} on every generator, whereas a \ac{GRU} memory layer helped the reactive baseline but added no benefit once subgoals were available.

Two design principles follow: (1)~generator diversity during training is essential for generalization, and (2)~coupling a global planner with a learned local policy outperforms both a reactive policy alone and a fixed-speed classical follower.

Future work can extend this in three directions: removing the dependence on $A^*$ via a hierarchical policy that learns its own subgoal stream end-to-end; lifting the single-robot setting to multi-robot fleets, where subgoal-style guidance becomes a natural coordination signal; and extending the sim-to-real evaluation to multi-robot intralogistics scenarios with dynamic obstacles and human co-workers.

\bibliographystyle{IEEEtran}
\bibliography{references}

\end{document}